\documentclass{article} % For LaTeX2e
\usepackage{iclr2023_conference_tinypaper,times}

\usepackage{amsmath,amsfonts,bm}

\def\eqref#1{equation~\ref{#1}}
\def\1{\bm{1}}

\DeclareMathAlphabet{\mathsfit}{\encodingdefault}{\sfdefault}{m}{sl}
\SetMathAlphabet{\mathsfit}{bold}{\encodingdefault}{\sfdefault}{bx}{n}

\usepackage{hyperref}
\usepackage{url}

\usepackage{graphicx}
\graphicspath{{./figure/}}
\usepackage{multirow}

\title{Prompt Engineering and Calibration for Zero-Shot Commonsense Reasoning}
\author{Chenkai Ma \\
School of Computer Science and Engineering \\
University of Electronic Science and Technology of China\\
Chengdu, 611731, China \\
\texttt{kasmas316@gmail.com}
}

\iclrfinalcopy % Uncomment for camera-ready version, but NOT for submission.
\begin{document}

\maketitle

\begin{abstract}

Prompt engineering and calibration make large language models excel at reasoning tasks, including multiple choice commonsense reasoning. From a practical perspective, we investigate and evaluate these strategies on smaller language models. Through experiments on five commonsense reasoning benchmarks, we find that each strategy favors certain models, but their joint effects are mostly negative.
% we find that prompt engineering benefits instruction-tuned models, while calibration brings mixed results. In addition, their joint effects tend to be negative.
\end{abstract}

\section{Introduction}
% One contribution: Exploring the combination of different strategies, which is not studied before. 
% Another contribution: Make modificaitons to these strategies to fit smaller lm.

Large Language models (LLMs) have shown impressive performance in many NLP applications (\citealp{Ouyang2022TrainingLM}; \citealp{Chung2022ScalingIL}; \citealp{wei2022finetuned}),
including commonsense reasoning, a key component to AGI \citep{Davis2015CommonsenseRA}. 
Recent studies suggest that LLMs are capable of zero-shot and few-shot learning (\citealp{NEURIPS2020_1457c0d6}; \citealp{webson-pavlick-2022-prompt}; \citealp{Chowdhery2022PaLMSL}), and that several strategies can further improve their performance, like prompt engineering and calibration (\citealp{kojima2022large}; \citealp{pmlr-v139-zhao21c}; \citealp{jiang-etal-2021-know}; \citealp{Kadavath2022LanguageM}). 
Despite achieving SOTA performance on many benchmarks, most LLMs are very expensive to use and not released to the public. 

Consequently, we study whether prompt engineering and calibration can help smaller language models (those with no more than 3B parameters) in zero-shot multiple choice commonsense reasoning. Since these strategies are likely emergent (\citealp{wei2022emergent}; \citealp{Chan2022DataDP}), we make several modifications, then evaluate them on five commonsense reasoning benchmarks.
We find that prompt engineering favors large Flan-T5 models, while calibration works well on GPT-2. Their joint effects are, however, negative in most cases.

% We find that prompt engineering are only helpful to instruction-tuned language models, while calibration brings mixed results. Moreover, the joint effect of these strategies are more likely to be negative.

\section{Methods}

\begin{figure}[ht]
    \begin{center}
    \includegraphics[width=1.0\linewidth]{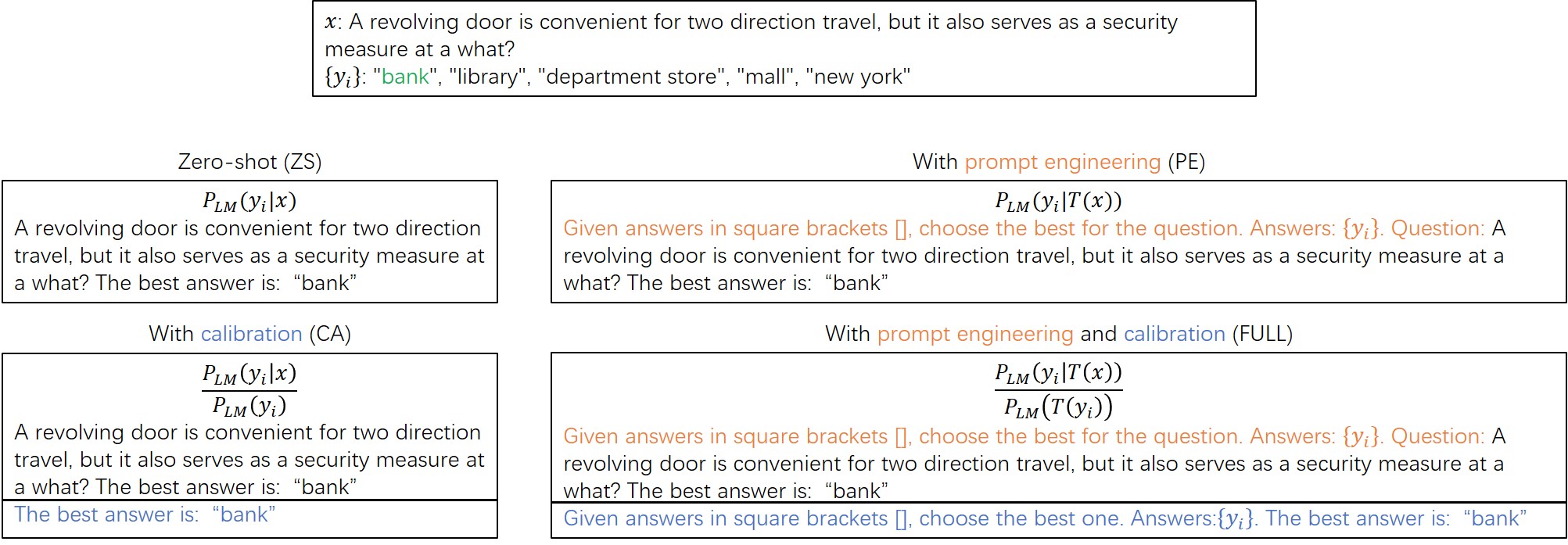} % scale=0.4
    \caption{Combinations of data format and option scores for multiple choice commonsense reasoning. Based on the zero-shot method, we add prompt engineering (instruction and multiple choice prompt) and calibration. Unlike previous works, we do not bind options to symbols, like (A).}
    \label{fig:method}
    \end{center}
\end{figure}

% \subsection{Background}
\textbf{Background.}
Multiple choice commonsense reasoning is formalized as follows: Given a question $x$ and several options ${y_{1},...,y_{n}}$, select the best option. In the zero-shot setting, a language model computes a score for each option, which is usually the conditional probability $P_{LM}(y_{i}|x)$, and selects the one with the highest score, as shown in Figure \ref{fig:method}. Recent works suggest that alternatives to the conditional probability can lead to better performance (\citealp{holtzman-etal-2021-surface}; \citealp{niu-etal-2021-semantic}; \citealp{min-etal-2022-noisy}), but we do not consider these variants for simplicity and fair comparison.

% Zero-shot methods use a language model to score each option, and choose the option with the highest score: $\hat{y}=\argmax_{i}score(y_{i})$. Typically, the score of an option $y_{i}$ is its conditional probability $P_{LM}(y_{i}|x)$ \textbf{ref}, while some alternatives have shown better performance \textbf{ref}.

% \subsection{Prompt Engineering: Multiple Choice Prompt and Instruction}
\textbf{Prompt engineering: multiple choice prompt and instruction.}
A limit of $P_{LM}(y_{i}|x)$ is that options are not considered jointly. Recent works suggest that providing all the options in the input, along with instructions about the task, can help LM reason (\citealp{robinson2023leveraging}; \citealp{Chung2022ScalingIL}). Inspired by these ideas, we design templates $T()$ that add an instruction and options to a question, as shown in Figure \ref{fig:method}. Unlike recent methods that bind each option to a symbol like (A), we use an LM to directly predict answers, because symbol binding is an emergent ability \citep{robinson2023leveraging}.

% Our method differs from those for LLMs in that ours predict direct answers, not symbols (e.g., (A), (B)) that represent answers. An example is show in Figure \ref{fig:method}.

% \subsection{Calibration}
\textbf{Calibration.}
Recent works find that language models prefer certain options even without a question, which suggests they are not well-calibrated (\citealp{pmlr-v139-zhao21c}; \citealp{jiang-etal-2021-know}). 
To overcome this problem, we divide the conditional score of an option by another score computed from a "null" prompt that contains no question, as in $\frac{P_{LM}(y_{i}|x)}{P_{LM}(y_i)}$. An example is shown in Figure \ref{fig:method}.

% Ideally, if we ask a language models to choose an answer from several options without any context, it should not have strong preference for any of them. Nevertheless, we find language models do have intrinsic preferences for some answers, which may be caused by pre-training. Therefore, we calibrate the language model by dividing its scores by its scores with no context.
% Another limit of $P_{LM}(y_{i}|x)$ is that these scores

\section{Experiments}

\textbf{Datasets.} We evaluate prompt engineering and calibration on five multiple choice commonsense benchmarks: 
% to cover different kinds of commonsense reasoning
(1) CommonsenseQA (CSQA) \citep{talmor-etal-2019-commonsenseqa}; (2) COPA \citep{gordon-etal-2012-semeval}; (3) OpenBookQA (OBQA) \citep{mihaylov-etal-2018-suit}; (4)PIQA \citep{Bisk2019PIQARA}; (5)Social IQA (SIQA) \citep{sap-etal-2019-social}; 
% (6)MC-TACO \citep{zhou-etal-2019-going}. 
We present their statistics in Appendix \ref{app:stat}. For all benchmarks, we only use their development sets.

\textbf{Baselines.} 
We compare four zero-shot methods mentioned in Figure \ref{fig:method}: (1) ZS, the standard zero-shot method that computes conditional probability scores of each option; (2) CA, which is ZS with calibration, also known as PMI\textsubscript{DC} in \citet{holtzman-etal-2021-surface}; (3) PE, which is ZS with prompt engineering; (4) FULL, which is ZS with both prompt engineering and calibration.

\textbf{Setup.} 
As for language models, we use GPT-2 \citep{Radford2019LanguageMA}, T5 \citep{10.5555/3455716.3455856}, and Flan-T5 \citep{Chung2022ScalingIL}, except Flan-T5-XXL, which is too large (11B) to store on our hardware. The evaluation metric is accuracy.

% \section{Results and Analysis}

\begin{table}[htbp]
    \caption{Accuracy (\%) on Flan-T5}
    \begin{center}
    \scalebox{0.60}{
    \begin{tabular}{c|c|c|c|c|c|c|c|c|c|c|c|c|c|c|c|c}
    \hline
        \multirow{2}{4em}{Model} & \multicolumn{4}{c}{Flan-T5-Small (80M)} & \multicolumn{4}{|c}{Flan-T5-Base (250M)} & \multicolumn{4}{|c}{Flan-T5-Large (780M)} & \multicolumn{4}{|c}{Flan-T5-XL (3B)} \\
        & ZS & CA & PE & FULL & ZS & CA & PE & FULL & ZS & CA & PE & FULL & ZS & CA & PE & FULL \\
    \hline  
        COPA & \textbf{59.8} & 56.6 & 52.0 & 49.6 & 67.0 & \textbf{68.2} & 60.6 & 61.4 & 72.8 & 71.6 & \textbf{87.6} & 84.0 & 80.8 & 78.4 & \textbf{88.8} & 85.6 \\ 
        CSQA & 29.2 & \textbf{37.7} & 30.8 & 28.3 & 40.9 & 48.5 & \textbf{52.5} & 51.8 & 51.6 & 51.5 & 62.2 & \textbf{67.6} & 61.8 & 64.7 & 70.6 & \textbf{72.7} \\
        OBQA & 14.0 & \textbf{32.6} & 24.8 & 29.6 & 20.0 & \textbf{34.0} & 28.6 & 34.0 & 24.2 & 39.4 & \textbf{53.4} & 52.8 & 30.0 & 49.6 & \textbf{61.0} & 55.4 \\
        PIQA & \textbf{62.5} & 57.6 & 54.2 & 51.1 & \textbf{65.9} & 59.7 & 58.1 & 54.0 & 71.4 & 65.5 & \textbf{72.7} & 60.6 & \textbf{75.8} & 68.3 & 68.9 & 60.4 \\
        SIQA & 41.7 & \textbf{42.5} & 42.3 & 42.3 & 46.4 & 47.4 & \textbf{54.7} & 53.7 & 51.4 & 48.1 & \textbf{68.6} & 66.7 & 56.1 & 56.3 & \textbf{71.6} & 58.9 \\
        % &  &  &  &  &  &  &  &  &  &  &  &  &  &  &  &  
    \hline
    \end{tabular}
    }
    \end{center}
    \label{tab:flant5}
\end{table}

% Findins: Separate findngs for Flan-t5 and other models.
\textbf{Results.}
We present results on Flan-T5 in Table \ref{tab:flant5}, and results on GPT-2 and T5 in Appendix \ref{app:other_lm}. 
We find prompt engineering does not work for most models, except the two largest Flan-T5 models, on which it boosts performance by as much as 30 points. This corroborates the effects of instruction-tuning on Flan-T5, and the emergent abilities of larger models \citep{Chung2022ScalingIL}.
Apart from that, calibration works well on GPT-2, but inconsistently on other models. This supports findings in \citet{holtzman-etal-2021-surface}.
Furthermore, the joint effects of both strategies are mostly negative. 
Overall, our findings suggest careful inspections when using these strategies, as there is no universal configuration that works well on all models.

% Accordingly, we find the strategies work differently on Flan-T5 and other two kinds of language models.
% For Flan-T5, prompt engineering does not work for the smallest model, but remarkably boosts performance for the other models, e.g., by 15 points on Flan-T5-Large on COPA. On the other hand, calibration has mixed effects on the models. In addition, when we use both strategies, their effects are also inconsistent.
% For GPT-2 and T5, calibration helps, improving over ZS as much as 20 percentage points. On the contrary, prompt engineering is mostly harmful, e.g., loses 18 percentage points on GPT-2 XL on COPA. When used together, the two strategies hurt performance in most cases.

% While Flan-T5 models are instruction tuned \citep{Chung2022ScalingIL}, the other two are not, which explains why they benefit from prompt engineering. Unlike previous works on calibration (\citealp{pmlr-v139-zhao21c}, \citealp{jiang-etal-2021-know}, \citealp{Kadavath2022LanguageM}), we find calibration brings mixed results, and suggest careful inspections of calibration to newer models like Flan-T5.

\section{Conclusion}
We study whether prompt engineering and calibration help smaller language models in multiple choice commonsense reasoning, as they help LLMs. We find that while each strategy favors some language models, their joint effects are mostly negative. Therefore, we suggest careful inspections of these strategies before applying them to smaller language models.
% We find leaningly negative joint effects from these strategies, among which prompt engineering only benefits instruction-tuned language models, and calibration brings mixed results. 

% For instruction-tuned models, prompt engineering is mostly beneficial, whereas calibration brings mixed results.

\section*{URM Statement}
Author Chenkai Ma meets the URM criteria of ICLR 2023 Tiny Papers Track.
% The authors acknowledge that at least one key author of this work meets the URM criteria of ICLR 2023 Tiny Papers Track.
% Please include this URM Statement section at the end of the paper but before the references before. In your anonymized submission, we recommend stating ``The authors acknowledge that at least one key author of this work meets the URM criteria of ICLR 2023 Tiny Papers Track.'' For the camera ready version, we ask authors to identify which author(s) meet the URM criteria, e.g., ``Author TFB meets the URM criteria of ICLR 2023 Tiny Papers Track.'' The authors are also welcome to come up with their own phrases to affirm meeting this criteria.

\bibliography{iclr2023_conference_tinypaper}
\bibliographystyle{iclr2023_conference_tinypaper}

\appendix

\section{Full Prompts for All Benchmarks} 
In this section, we present prompts (i.e., templates) for each benchmark in Table \ref{tab:prompts}. Specifically, we use one prompt for CSQA and SIQA, and another for COPA, OBQA, and PIQA, because the latter three do not always have a question in a data sample. For simplicity, we still use the term "question" for these three datasets. We also provide the prompts we use for calibration, which is used in FULL.

\begin{table}[htbp]
    \caption{Prompts for each benchmark}
    \begin{center}
    \scalebox{0.7}{
    \begin{tabular}{c|c|c}
    \hline
        Benchmarks & Prompt for the Question & Prompt for Calibration \\
    \hline
       \multirow{3}{*}{CSQA, SIQA} & \multirow{3}{22em}{Given answers in square brackets [], choose the best for the question. Answers: [\textit{answers}]. Question: [\textit{question}] The best answer is: } & \multirow{3}{20em}{Given answers in square brackets [], choose the best one. Answers: [\textit{answers}]. The best answer is: } \\
         & & \\ 
         & & \\
         \hline
        \multirow{3}{*}{COPA, OBQA, PIQA} & \multirow{3}{22em}{Given answers in square brackets [], choose the one that best completes the sentence. Answers: [\textit{answers}]. Sentence: [\textit{question}] The best answer is: } & \multirow{3}{20em}{Given answers in square brackets [], choose the best one. Answers: [\textit{answers}]. The best answer is: } \\ 
         & & \\
         & & \\
         \hline
    \end{tabular}
    }
    \end{center}
    \label{tab:prompts}
\end{table}

\section{Dataset Statistics}
\label{app:stat}
We present statistics of the five commonsense reasoning (CSR) dataset we use in our experiments in Table \ref{tab:stat}.
\begin{table}[htbp]
    \caption{Statistics of datasets}
    \begin{center}
    \scalebox{0.8}
    {
    \begin{tabular}{c|c|c|c|c|c}
    \hline
     Dataset Name& Type of CSR & Number of choices & Train & Validation & Test  \\
    \hline
     COPA \citep{gordon-etal-2012-semeval} & Causal & 2 & N/A & 500 & 500 \\
     CSQA \citep{talmor-etal-2019-commonsenseqa} & General & 5 & 9741 & 1221 & 1140 \\
     % MC-TACO \citep{zhou-etal-2019-going} & Temporal & TBD & N/A & 561 & 1332 \\
     OBQA \citep{mihaylov-etal-2018-suit} & Scientific & 4 & 4957 & 500 & 500 \\
     PIQA \citep{Bisk2019PIQARA} & Physical & 2 & 16000 & 2000 & 3000 \\
     SIQA \citep{sap-etal-2019-social} & Social & 3 & 33410 & 1954 & N/A \\
    \hline
    \end{tabular}
    }
    \end{center}
    \label{tab:stat}
\end{table}

\section{Results on GPT-2 and T5}
\label{app:other_lm}
We present results on GPT-2 in Table \ref{tab:gpt2}, and T5 in Table \ref{tab:t5}.
\begin{table}[htbp]
    \caption{Accuracy (\%) on GPT-2}
    \begin{center}
    \scalebox{0.60}{
    \begin{tabular}{c|c|c|c|c|c|c|c|c|c|c|c|c|c|c|c|c}
    \hline
        \multirow{2}{4em}{Model} & \multicolumn{4}{c}{GPT-2-Base (125M)} & \multicolumn{4}{|c}{GPT-2-Medium (350M)} & \multicolumn{4}{|c}{GPT-2-Large (765M)} & \multicolumn{4}{|c}{GPT-2-XL (1.6B)} \\
        & ZS & CA & PE & FULL & ZS & CA & PE & FULL & ZS & CA & PE & FULL & ZS & CA & PE & FULL \\
    \hline  
        COPA & 61.0 & \textbf{62.8} & 53.0 & 54.4 & 67.0 & \textbf{70.0} & 49.4 & 54.2 & \textbf{69.8} & 69.4 & 51.4 & 57.4 & 69.0 & \textbf{71.6} & 51.4 & 53.0 \\ 
        CSQA & 25.5 & \textbf{36.4} & 23.8 & 27.4 & 30.9 & \textbf{41.8} & 27.4 & 30.1 & 33.3 & \textbf{44.5} & 26.9 & 33.2 & 38.6 & \textbf{47.8} & 35.1 & 36.2 \\
        OBQA & 15.8 & \textbf{33.4} & 25.6 & 28.0 & 18.0 & \textbf{38.6} & 26.8 & 27.4 & 21.6 & \textbf{41.4} & 25.2 & 29.4 & 22.4 & \textbf{43.2} & 25.8 & 29.4 \\
        PIQA & \textbf{62.1} & 57.1 & 54.6 & 52.6 & \textbf{66.2} & 57.5 & 51.8 & 52.6 & \textbf{69.6} & 60.7 & 55.0 & 54.6 & \textbf{69.6} & 62.2 & 52.6 & 53.4 \\
        SIQA & 35.8 & \textbf{38.0} & 34.3 & 37.1 & 36.9 & \textbf{40.0} & 36.0 & 38.0 & 36.6 & \textbf{40.3} & 34.0 & 35.6 & 39.0 & \textbf{41.0} & 35.2 & 35.9 \\
        % &  &  &  &  &  &  &  &  &  &  &  &  &  &  &  &  
    \hline
    \end{tabular}
    }
    \end{center}
    \label{tab:gpt2}
\end{table}

\begin{table}[!ht]
    \caption{Accuracy (\%) on T5}
    \begin{center}
    \scalebox{0.60}{
    \begin{tabular}{c|c|c|c|c|c|c|c|c|c|c|c|c}
    \hline
        \multirow{2}{4em}{Model} & \multicolumn{4}{c}{T5-Small (80M)} & \multicolumn{4}{|c}{T5-Base (250M)} & \multicolumn{4}{|c}{T5-Large (780M)} \\
        & ZS & CA & PE & FULL & ZS & CA & PE & FULL & ZS & CA & PE & FULL \\
    \hline  
        COPA & \textbf{55.2} & 51.2 & 51.2 & 52.2 & \textbf{59.6} & 59.4 & 51.0 & 51.8 & \textbf{65.2} & 56.6 & 53.2 & 53.8 \\ 
        CSQA & 16.6 & \textbf{22.8} & 21.1 & 21.0 & 26.1 & \textbf{30.0} & 20.6 & 22.5 & \textbf{39.2} & 35.4 & 33.1 & 35.7 \\
        OBQA & 14.2 & \textbf{28.8} & 23.8 & 25.8 & 15.8 & \textbf{30.8} & 27.8 & 27.2 & 19.0 & \textbf{30.4} & 24.8 & 26.4 \\
        PIQA & \textbf{56.6} & 50.5 & 51.2 & 50.8 & \textbf{61.0} & 57.7 & 51.7 & 53.0 & \textbf{66.6} & 64.4 & 52.8 & 51.7 \\
        SIQA & \textbf{36.2} & 36.1 & 35.0 & 34.4 & 36.2 & \textbf{37.6} & 37.0 & 33.5 & \textbf{38.7} & 38.1 & 37.0 & 34.1 \\
        % &  &  &  &  &  &  &  &  &  &  &  &  &  &  &  &  
    \hline
    \end{tabular}
    }
    \end{center}
    \label{tab:t5}
\end{table}

\section{Code}
\label{app:code}
Our code is available at \url{https://anonymous.4open.science/r/Prompt-engineering-and-calibration-0AE0/README.md}

\end{document}